\documentclass[a4paper, 10 pt, conference]{IEEEtran}
\usepackage{eurosym}
\usepackage{upgreek}
\usepackage[left=1.9cm, right=1.3cm, bottom=2.1cm, top=3.66cm]{geometry}

\usepackage{cite}

\ifCLASSINFOpdf
  \usepackage[pdftex]{graphicx}
  \usepackage{import}

\usepackage[english]{babel} 
\usepackage[utf8]{inputenc}
\usepackage[]{caption}
\usepackage{subcaption}
\usepackage{placeins}
\usepackage{color}
\usepackage{mathtools}
\usepackage{placeins}
\usepackage{relsize}
\usepackage{amssymb}
\usepackage{units}
\usepackage{amsmath}
\usepackage{float} 		
\renewcommand{\figurename}{Fig.}
\addto\captionsenglish{\renewcommand{\figurename}{Fig.}}
\usepackage{nicefrac}
\usepackage{tikz}
\usepackage[percent]{overpic}

\definecolor{mygray}{RGB}{70, 70, 70}
\definecolor{mygray2}{RGB}{40, 40, 40}

\usepackage{pgf}
\usepackage{tikz}
\usetikzlibrary{arrows,automata}
\usetikzlibrary{positioning}
\tikzset{    state2/.style={ rectangle,  draw=black, inner sep=7pt,  text centered  },}
\usepackage[color=black,opacity=1,angle=0,scale=1]{background}
\backgroundsetup{
contents={
\begin{tikzpicture}
\node at (current page.center) [align=center] {\textcopyright 2020 IEEE. Personal use of this material is permitted. 
\\ Permission from IEEE must be obtained for all other uses, in any current or future media, including reprinting/republishing this material for advertising or promotional purposes, 
\\ creating new collective works, for resale or redistribution to servers or lists, or reuse of any copyrighted component of this work in other works.
\\ DOI: 10.1109/ITSC45102.2020.9294452};
\end{tikzpicture}},
placement=bottom,
scale=0.6,
vshift=20
}

\begin{document}
\title{\vspace*{0.4cm} 
Proactive Risk Navigation System for Real-World Urban Intersections
\vspace*{-0.3cm}} 
\author{\IEEEauthorblockN{Tim Puphal$^{1\star}$, Benedict Flade$^{1\star}$, Daan de Geus$^{2}$ and Julian Eggert$^{1}$} 
\IEEEauthorblockA{$^{1}$ Honda Research Institute (HRI) Europe, 
	Carl-Legien-Str. 30, 63073 Offenbach, Germany \\
	Email: {\tt\small \{tim.puphal, benedict.flade, julian.eggert\}@honda-ri.de} \\
	$^{2}$ Eindhoven University of Technology (TU/e), 5612 AZ Eindhoven, the Netherlands \\
	Email: {\tt\small d.c.d.geus@tue.nl}} 
    $\star$ The authors contributed equally to this work}
    
\maketitle

\begin{abstract} 
We consider the problem of intelligently navigating through complex traffic. Urban situations are defined by the underlying map structure and special regulatory objects of e.g. a stop line or crosswalk. Thereon dynamic vehicles (cars, bicycles, etc.) move forward, while trying to keep accident risks low. 

Especially at intersections, the combination and interaction of traffic elements is diverse and human drivers need to focus on specific elements which are critical for their behavior.
To support the analysis, we present in this paper the so-called Risk Navigation System (RNS). RNS leverages a graph-based local dynamic map with Time-To-X indicators for extracting upcoming sharp curves, intersection zones and possible vehicle-to-object collision points. 

In real car recordings, recommended velocity profiles to avoid risks are visualized within a 2D environment. 
By focusing on communicating not only the positional but also the temporal relation, RNS potentially helps to enhance awareness and prediction capabilities of the user. 
\end{abstract}

\IEEEpeerreviewmaketitle

\vspace{0.16cm}
\section{Introduction}
\subsection{Motivation}
Human error has been proven to be one of the main causes of accidents in public driving with a share of over $\unit[90]{\%}$. The remaining $\unit[10]{\%}$ of the crashes happen because of technical vehicle failure or bad weather conditions. According to the World Health Organization (WHO)~\cite{who2004}, drivers encounter particularly difficult situations due to:
\begin{itemize} 
 \item Inexperience, recklessnes and fatigue which leads to violating the recommended speed or having inappropriate distances to the roadside and other cars.
 \item A limited field of view (sometimes caused by environmental objects, i.e., trees or buildings) which results in failing to see the point of interest. The reaction time for suddenly appearing obstacles can thus be critical. 
 \item Incapability to estimate the behavior of other participants. The deviation of the real situation and resulting ego action may generate unavoidable collisions.
\end{itemize}

During recent years, Advanced Driver Assistance Systems (ADAS) were developed to take over certain tasks in the control of the car. In total, about $\unit[33]{\%}$ of the occuring crashes represent single-vehicle events, $\unit[27]{\%}$ often correspond to longitudinal following or lane changes and $\unit[40]{\%}$ occur in intersection areas~\cite{nhtsa2008}.  
ADAS can already succesfully tackle the first and second category events with e.g. anti-lock braking systems, electronic stability control, adaptive cruise control and lane keeping assistance. However for general intersections, solutions are limited and no holistic ADAS is comercially available yet. Often, collision mitigation systems~\cite{bengler2014} simply execute an emergency braking shortly before an obstacle is crossing. More proactive ADAS are needed to prepare the driver in advance for different hazards arising from urban traffic.  

For this purpose, we present the so-called Risk Navigation System (RNS). RNS is based on a local dynamic map that serves as a central hub for road data and sensor measurements. We reduce the complexity of intersection scenarios by extracting relevant possible paths within a virtual horizon, regulatory objects (e.g. stop line, crosswalk and traffic light) as well as vehicle interaction zones (like junction areas and point of closest encounters with other cars). 
Consequently, Time-To-X indicators are employed to detect critical future situations and to recommend a safe behavior. 

\begin{figure}
  \centering
  \resizebox{\linewidth}{!}{
  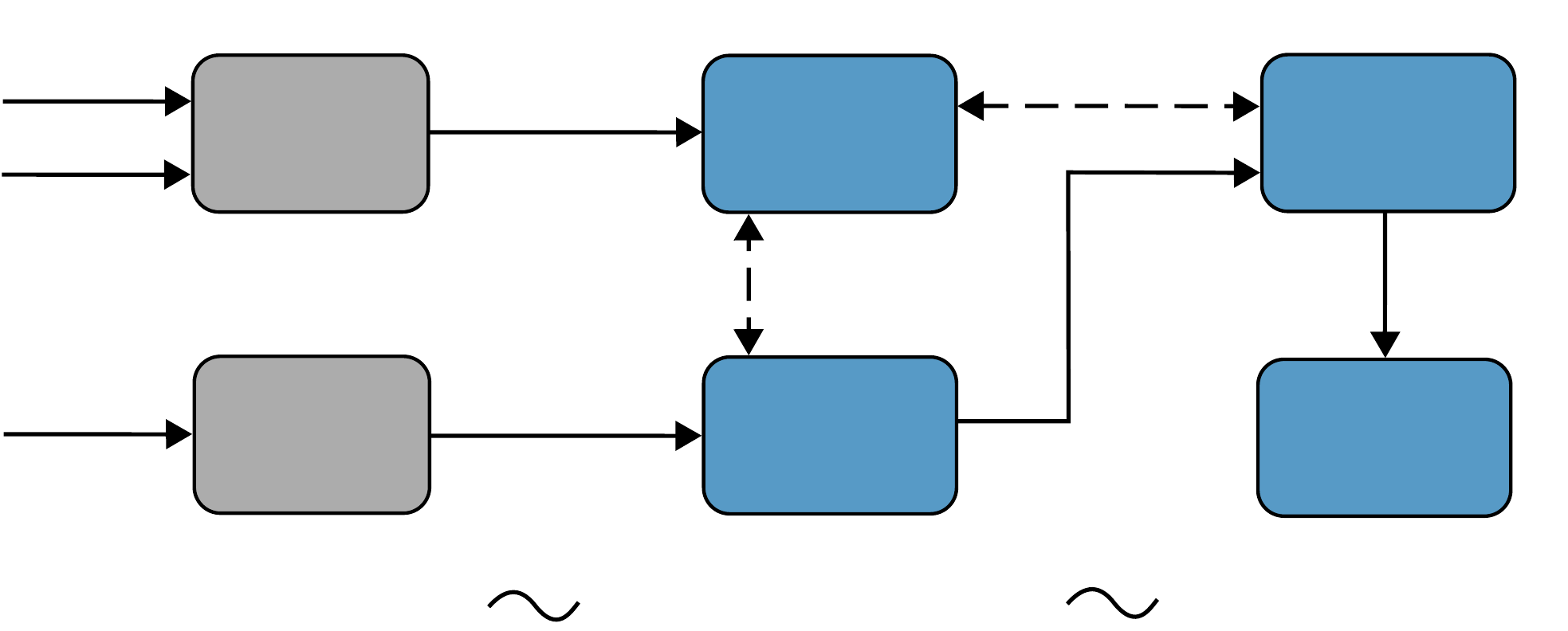}
  \caption{Concept of the Risk Navigation System (RNS), designed for driver support at general intersections.}
  \vspace*{-0.15cm}
  \label{fig:blockdiagram} 
\end{figure}

If the user does not behave in a risk-appropriate way, RNS displays warnings that convey both the positional and temporal information related to the encountered risks. In real-time recordings combining GNSS\footnote{Accepted term for Global Navigation Satellite System (GPS, Galileo, GLONASS, etc.).} and lidar, we show that RNS is effective in analyzing and visualizing collision, curve and regulatory risk. Awareness and prediction abilities of the driver are improved. In Figure \ref{fig:blockdiagram}, the pipeline of the realized system is given. In this paper, our focus with RNS lies on the boxes indicated in blue.

The remainder of the paper is structured as follows. First, related scientific work for risk warning is summarized with section \ref{sec:rel}. Afterwards, we explain our methods in detail in sections \ref{sec:riskextr} and \ref{sec:visrisk}. Central addressed topics are the storage, filtering and visualization of risk information using the local dynamic map. With section \ref{sec:fite}, the application of RNS and qualitative outputs are shown. Finally, we give a conclusion and outlook for future work in section \ref{sec:outl}.

\subsection{Related Work}
\label{sec:rel}
Communication of risks is a complex task. Traffic situations have high uncertainty and variability of how they might develop. 
On the one hand, a proper Human-Machine Interface (HMI) is required to provide sufficient context information around the ego vehicle (e.g. risk source or collision severity). On the other hand, driving risks lie in the range of seconds and need to be understood in a fast way. The HMI should be clear, intuitive and easily interpretable~\cite{eppler2008}. 
According to \cite{knoll2007}, the underlying three steps for associated warning systems consist of a) retrieval of risk indicators from sensors and maps, b) comparison of actual versus planned behavior and c) on demand application of HMI modalities in visual, audio or tactile form. 

Traffic risk extraction is commonly achieved with the help of outside sensors (camera, lidar and radar). Exemplarily, the authors of~\cite{messelodi2008} identify construction signs and vehicles in a front-mounted camera. For headway control or crash warning, \cite{gat2005} compute thereof distance and Time-To-Collision (TTC). Fusing camera with radar allows the reduction of noise in the object velocity estimation. 
Both approaches act in a relative coordinate system, while all inputs are received on-the-fly. 
In contrast, ego sensors (e.g. a GNSS plus intertial measurement unit) can be aligned in a global reference frame for leveraging map data. A real-time setup is shown in~\cite{nexyad2019}. Their HMI displays pop-up symbols from tagged speed limits or occluded crosswalks. Similarly, \cite{pu2015} employ road-level but calculate the curvature of the street for saving up fuel via an engine management. 

The RNS as proposed in this work relies on a relational local dynamic map that concurrently combines GNSS with lidar sensors. Effective retrieval of relevant risk sources from paths, traffic signs or vehicles is ensured with a graph instead of table database. For more literature about local dynamic maps, risk estimation or motion planning, please refer to \cite{ldm2017}, \cite{puphal2019} and \cite{puphal2018}. 

While most ADAS apply various visual monitors, audible and tactile actuators are still a topic of research. In \cite{alves2008}, a head-up display is enhanced by sound to shorten the response time of drivers. On top of that, \cite{dettmann2016} investigate light strips for spatially distant information and alarms for immanent critical entities. Concerning navigation devices, tactile impulses on a car seat~\cite{chang2011} can be applied in turning maneuvers, or mobile belts~\cite{krueger2018} vibrate according to the relative position of obstacles. In the process, the priority of elements is assigned by varying the press intensity. 

By comparison, visual modality supports animation elements and color coding for conveying urgency. Nevertheless, in handovers between machine and human, situation understanding is crucial and visual context is necessary. This is based on the circumstance that humans mainly rely on visual perception during driving~\cite{plavsic2009}. As a reference, typical HMI transitions in a single-risk scenario last on average 2 to 8 seconds~\cite{walch2015}. 

Symbolic displays in the cockpit (warning with text or icons) have been thoroughly investigated in user studies, e.g. compare~\cite{plavsic2009}. It turned out that workload and effort may increase with a 3D cluster opposed to 2D traffic environment. Bird's-eye views are additionally rated as rather clear and comfortable than first person views.
In order to provide a $\unit[360]{^{\circ}}$ analysis, \cite{winner2016} lately introduced safety rings which bend to indicate positions of dangerous vehicles.  
Analogically, lateral lane change plus longitudinal speed recommendation can be conveyed by visual HMIs \cite{habenicht2011}. Lastly, the field of augmented reality is gaining more and more attention. In order to intuitively display vehicle distance and velocity, \cite{roessing2013} artificially blur the motion and create depth on high-quality videos. Further tests were done in \cite{narzt2004} where colored lanes are embedded into images, e.g. right turns as green, left turns as red and straight crossing as yellow. 

For our RNS, we choose to employ multiple visual elements of charts, pop-up symbols and animations inside a 2D birds-eye-view from the ego surrounding. The simulator naturally supports 3D views in other perspectives as well. In summary, RNS navigates the user along a route to an urban goal safely across intersections by issuing predictive informations, recommendations as well as warnings.

\section{Map-Based Risk Prediction}
\label{sec:riskextr}
The target is to represent driving situations with a flexible and efficient environment model. Generalized ADAS for intersections require clear data management functions.  
The following section \ref{subsec:ldm} shows how a graph-based local dynamic map can be leveraged for this purpose. Additionally, risk-relevant data within a certain virtual horizon is retrieved to allow predictive trajectory evaluations and planning in section \ref{subsec:trajeval}. 

\subsection{Relational Local Dynamic Map (R-LDM)}
\label{subsec:ldm}
Underlying road geometry significantly influences and constrains all traffic participants. 
From the ego driver, multi-faceted sensory inputs have to be processed in real-time.
Since important meta data lies in the connectivity of environment entities,
a graph representation, consiting of nodes, attributes and relations, is beneficial. 
The Relational Local Dynamic Map (R-LDM) presents an abstraction layer between sensor data and higher level functions~\cite{ldm2017}.
Such functions include e.g. risk prediction and visualization as required by our RNS. 
With the R-LDM, we support the paradigm shift from single sensor-controller loops towards technologies with enhanced understanding.

\subsubsection{Graph Structure}

A R-LDM graph stores data on four layers based on their dynamicity (compare Fig. \ref{fig:rldm} top right).
The map is represented as nodes with additional data of shape, orientation or type saved as attributes. 
On the lowest layer, the street geometry presents one of the most crucial data types. Generally, \textbf{static data} is stored in three levels of granularity, ranging from detailed lanes via half roads to the whole road.
In this context, half roads are the sum of all lanes pointing in the same direction.
Furthermore, the road subgraph consists of alternating segments and junctions, while all junctions at one location forms an intersection.

The second layer comprises \textbf{quasi-static data} with changes on the scale of several days. 
This implies traffic lights, roadside infrastructure, traffic signs or construction sites. 
On the third layer, \textbf{transient data} is stored changing on hours scale or faster (i.e., traffic light states, car density or slippery road conditions). 
Finally, the fourth layer contains highly \textbf{dynamic data} about the state of traffic participants, including vehicles and pedestrians. 
Detected dynamic objects are connected to specific nodes of the lower layers, which can be used for risk extraction in the ego vicinity.
The R-LDM is consequently maintained or updated in real-time.

With regard to lower layers, available map data sources such as the crowd-sourced OpenStreetMap (OSM)\footnote{See www.openstreetmap.org for further explanation.} can be applied. 
OSM mainly consists of navigational information represented in the form of road centerlines and intersection points. 
Therefore, we add topological information on lane level and estimate the actual lane geometry under assumption of a concrete lane width. 
Besides, stop lines are manually added and linked to incoming lanes, while the positions of traffic lights, crosswalks or buildings are similarly extracted from OSM. 

\vspace{0.06cm}
\subsubsection{Virtual Horizon}
\label{sub:horizon}

The graph structure with its connected nodes suits itself to a wide scope of potential queries with regard to path finding, routing or tree search. 
In this way, extraction of behaviorally relevant situations is possible in a straightforward way.
One crucial information for risk estimation is the knowledge of the upcoming driving path of traffic participants.
When considering start and end points in the map, we can intuitively query a \textit{sequence of nodes} (segments and junctions) along the route by traversing its relations. 

\begin{figure}
  \centering
  \resizebox{\linewidth}{!}{
  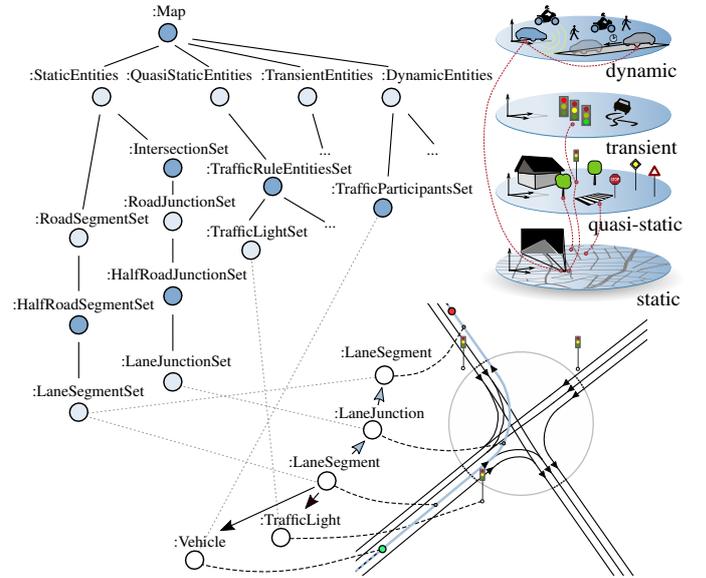}
  \caption{Nodes hierarchy and interconnection graph structure within R-LDM for an intersection with traffic lights.}
  \label{fig:rldm}
\end{figure}

Fig. \ref{fig:rldm} shows an example of an intersection scenario. 
Therein, we obtain lane nodes (:LaneSegment, :LaneJunction) and traffic light nodes (:TrafficLight), while the graph is explored based on the current position and orientation of a chosen vehicle (:Vehicle). 
Relevant data is stored in the attributes of the individual nodes. 
Essentially, we can query any processed information that has been acquired online (e.g. from sensor data) or offline (e.g. from OSM).

The left part of Fig. \ref{fig:rldm} depicts the corresponding excerpt of the graph concept that highlights layers (:StaticEntities, :QuasiStaticEntities, :TransientEntities, :DynamicEntities) and granularity.
In order to successfully calculate risk, not only one but several \textit{possible paths} have to be queried. 
In particular, a virtual horizon can be determined which extracts predicted paths for the vehicles in the traffic scene.

\vspace{-0.05cm}

\begin{figure*}
  \centering
  \resizebox{0.82\linewidth}{!}{
  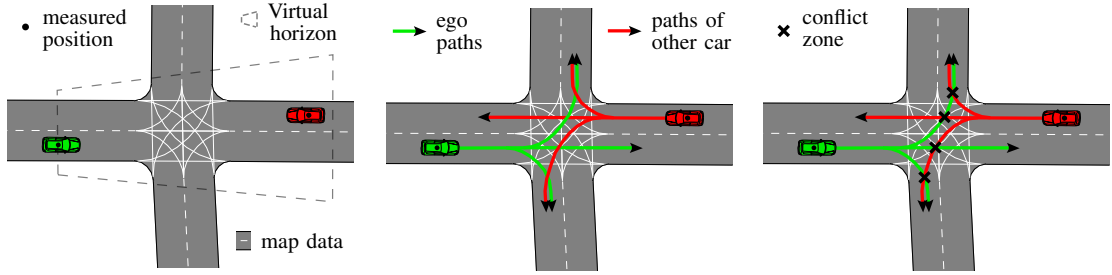}
  \caption{Steps of the virtual horizon shown exemplarily for collision risk. Left: Filtered relevant cars and their positions. Middle: Extraction of possible future paths. Right: Map-based risk extraction, with crosses indicating future path intersections.}
  \label{fig:virtualhorizon} 
\end{figure*}

\vspace{0.05cm}
To illustrate this virtual horizon, Fig. \ref{fig:virtualhorizon} addresses a scenario with two approaching cars. 
First, we map the GNSS-based ego vehicle position (in green) onto the center of the corresponding lane element.
Afterwards, we retrieve other cars from lidar (in red) up to a certain distance with R-LDM queries. More stored elements, for instance, quasi-static data with crosswalks can be equally obtained.

By scanning lane nodes for all cars, concatenating their polylines and ordering the branches (straight, left and right, 
etc.) according to a \textit{tree structure}, we obtain distinct paths. In Fig. \ref{fig:virtualhorizon}, every car posseses three possible paths.
The root of the tree is the actual position and the length increases with the number of segments we traverse.
With the virtual horizon, we can directly estimate simple risks. Crossings of the paths reflect conflict zones, where accidents might happen. These are unique and critical for the given road.

\subsection{Trajectory Evaluation}
\label{subsec:trajeval}
After querying the R-LDM for upcoming map structures, traffic elements and sensed objects, we predict trajectories from the vehicle dynamics and evaluate their concrete risk time-courses. Proper safe maneuvers or velocities for recommendation are related to our behavior planning. We divide risks for this purpose into three generic types: 1) collision risk, 2) curvature risk and 3) regulatory risk. In RNS, we combine the risk types to holistically support the driver.

\begin{figure}
  \centering
  \resizebox{0.80\linewidth}{!}{
  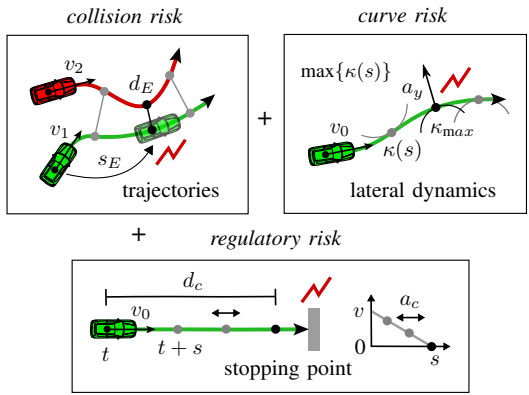}
  \vspace{0.08cm}
  \caption{Three risk types of RNS with time as well as space metrics (collision, curve and regulatory risks).}
  \label{fig:risktypes} 
\end{figure} 

\vspace{0.06cm}
\subsubsection{Collision Risk} 
Besides coarse conflict zones from the last section \ref{subsec:ldm}, risk metrics of quantitative collision probability can be obtained as well.
In an initial step, paths are transformed with constant velocity models into trajectories. For a car-to-car encounter (indexed 1,2), this gives us position sequences $\mathbf{x}_1(s)$ and $\mathbf{x}_2(s)$ over the predicted time $s$ with assumed constant velocities $v_1$ and $v_2$. 

The distance between the vehicles represents the pointwise trajectory difference and can be written as 
\begin{equation}
d(s)=\left\| \mathbf{x}_2(s) - \mathbf{x}_1(s) \right\|.
\end{equation}
For time risks, we next consider the time of maximal criticality along the predicted trajectories. In other words, we filter out the minimal distance, which is 
the event distance $d_E$ and get the corresponding Time-To-Closest-Encounter (TTCE) as $s_E$. The two indicators are in summary retrieved with

\begin{equation}
d_E=\text{min}\{d(s)\} \text{ and } s_E=\text{argmin}\{d(s)\}.
\end{equation}
\vspace{0.05cm}
\hspace{-0.114cm}Coarsely, collision risk corresponds to the inverse $1/s_E$ and is only considered if $d_E \hspace{-0.02cm} < \hspace{-0.02cm} d_{\text{min}}$ holds true. The minimal value $d_{\text{min}}$ defines the risk sensitivity. 

Once there are multiple trajectory pairs as in Fig. \ref{fig:virtualhorizon}, we would iterate over the combinations $\text{\textbf{x}}_{ij}$ (i.e., variation of ego paths $i$ and other paths $j$) and extract the highest risk by $\text{max}\{1/s_{E,ij}\}$. In this method, we therefore capture time and space of passing vehicles. 

\subsubsection{Curvature Risk}
From quantitative risks, we can also infer behavior recommendations as target velocities that minimize criticality and maximize ego utility. This is essentially a harder problem, since the driver should now understand the warning as well as interpret and process the advice. 

Consider the situation that the ego user is driving with velocity $v_0$ at the current time $t$. We define a turn segment with curvatures $[\kappa_{\text{start}}, \kappa_{\text{end}}]$ in the extracted path, if a thresholding condition
\begin{equation}
\kappa_{\text{start}} > \kappa_{\text{th}} \text{ and } \kappa_{\text{end}} < \kappa_{\text{th}}
\end{equation}
is valid. Here, the threshold $\kappa_{\text{th}}$, similarly as $d_{\text{min}}$, lets us adjust the turn detection along the future profile $\kappa(s)$. 

In curves, the vehicle is prone to drive off the lane. For higher velocities $v_0$, increased lateral acceleration is exerted. When $a_y$ depicts the dynamics limit that the vehicle can follow and $\kappa_{\text{max}} \hspace{-0.07cm} = \hspace{-0.07cm} \text{max}\{\kappa(s)\}$ is the maximal curvature in the segment, we can derive a target velocity via
\begin{equation}
v_{\text{tar}} = \sqrt{a_y/\kappa_{\text{max}}}.
\end{equation}
The variable $v_{\text{tar}}$ indicates the speed that should be maximally reached at the curve.  For low risk, we should accordingly move slower than $v_{\text{tar}}$, which leads to the condition $v_0 < v_{\text{tar}}$. Driving with $v_0 \ll v_{\text{tar}}$ is also not recommended, because of the reduction in utility.


\subsubsection{Regulatory Risk}
Quasi-static elements of a stop line, traffic light or crosswalk create risks at rule violation. The reason behind this circumstance is that generally not obeying the norms can lead to unexpected situations with accidents. Only because the vehicles are rule-conform, we can drive safely. 

In this sense, we may want to stop directly in front of the traffic data. Mathematically, we assume a soft braking trajectory with constant deceleration $a_c$ to the stopping point in the distance $d_c$ along the ego path. With the time $s$, we initially obtain the basic kinematic equation
\begin{equation}
d = \frac{a \cdot s^2}{2}.
\label{eq:distancestop}
\end{equation}
Substituting $s$ with $v_{\text{tar}}/a_c$ and inserting $a=a_c$ as well as $d=d_c$ in Eq. (\ref{eq:distancestop}) ultimately results into
\begin{equation}
v_{\text{tar}} = \sqrt{2a_c \cdot d_c},
\label{eq:vtar}
\end{equation}
whereby $v_{\text{tar}}$ is the target velocity to be able to stop at the intersection with certainty.\footnote{For Eq. (\ref{eq:vtar}), we do not account for the reaction time $t_r$. However, $t_r$ can be added for large distances $d_c$ with $v_{\text{tar}} \approx \sqrt{2a_c \cdot d_c} - a_c\cdot t_r$.} 

Fig. \ref{fig:risktypes} illustrates the corresponding variables for the three types of risk from this section. We want to stress at this point that our approach employs explicit trajectory prediction along the map paths from the R-LDM. Concretely, we assume constant velocity for the trajectory prediction in the collision case, an acceleration or deceleration trajectory to a fixed reduced velocity in the curve case and a smooth braking trajectory to a fixed position for the regulatory case. Therewith, we eventually filter and consider single time points in the trajectory for the risks.

\section{Visualization On Demand} 
\label{sec:visrisk}
Based on environment data and trajectory evaluation, we now present ways of communicating the situation and risks on a visual display to achieve an ADAS.
In this context, we employ a renderer that visualizes all the information in a joint Cartesian coordinate system (see section \ref{subsec:sim}). 
Once driving risks are detected, design elements are overlayed on the display with section \ref{subsec:active} and section \ref{subsec:warning}. 

\subsection{Simulator Environment}
\label{subsec:sim}
\begin{figure}[t]
  \centering
  \includegraphics[width=0.95\linewidth]{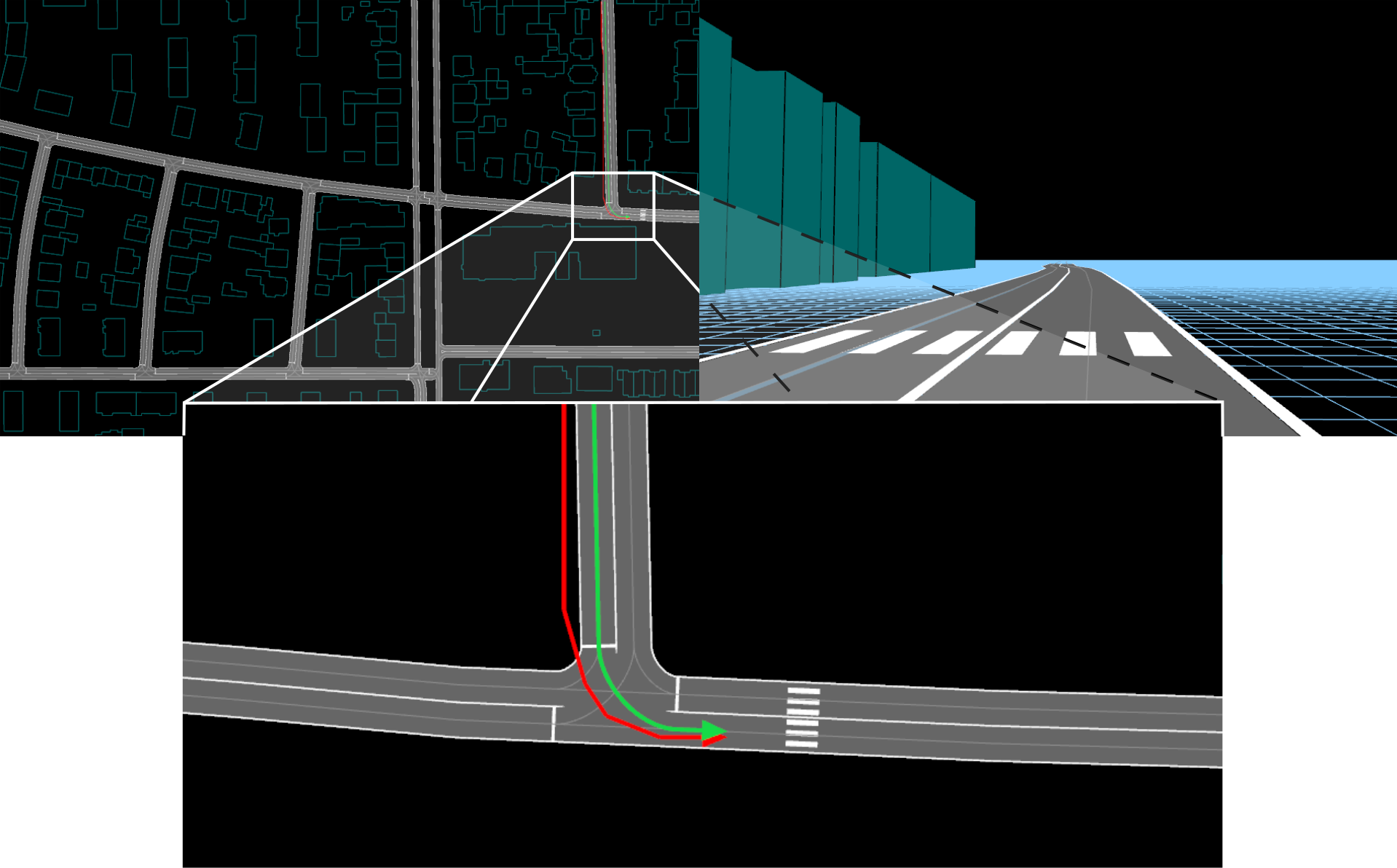}
  \caption{Rendered road network from two perspectives with the ego position being projected on the navigation route. \vspace{0.45cm}}
  \label{fig:3Dsimulator}
\end{figure}

Nodes of the R-LDM have a range of potential attributes, such as the 3D position or geometrical shape of objects. 
In the renderer, we always visualize static and quasi-static data that lie in the field of view from the ego vehicle. 
For this, a local 3D model is generated by converting geographic points with (lat, lon, alt) into Cartesian coordinates of (x, y, z). 
Fig. \ref{fig:3Dsimulator} depicts an exemplary map section having several intersections in bird's-eye view.
On the top right, the first person view of a vehicle approaching a crosswalk is shown. 

The dynamic data is then added to this static view. A zoomed-in excerpt from the map is given at the bottom of Fig. \ref{fig:3Dsimulator} that includes a recorded GNSS trace (red).
We project the trace onto the connected lane center, which is pictured in green. 
Consequently, the virtual horizon and its possible paths are retrieved as described in section \ref{subsec:ldm}. 
We can lastly update and move the excerpt with the current position from the GNSS to obtain a live simulation.

\subsection{Proactive Support}
\label{subsec:active}
Communication of spatial as well as spatio-temporal relations is crucial for risk-averse driver support. 
Further sources of information are cause, likelihood and severity of a potential risks.  
The next step for RNS is the choice of suitable design elements. 
In this process, we suppose that we know where the ego vehicle is driving (i.e., the ego path) from its navigation route. 
Yet, for surrounding vehicles, all paths are considered.

\subsubsection{Hazard Route Element}
\begin{figure}[t]
  \centering
  \resizebox{\linewidth}{!}{
  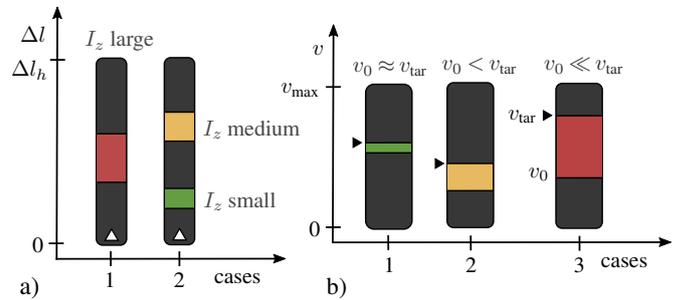}  
  \caption{Chart elements for proactive support. Hazard route (left) and velocity scale (right).} 
  \label{fig:charts} 
\end{figure}

The so-called hazard route in Fig. \ref{fig:charts} is a concept that consists of a scale portraying distances to an upcoming risk element.
Furthermore, the geometrical area or length of risks is considered.
Risk is thus measured with respect to the ego path, ranging from the current position  $\Delta l \hspace{-0.03cm}=\hspace{-0.03cm} \unit[0]{m}$ to the end of the path $\Delta l_{h}$.
Here, the length $\Delta l_{h}$ can be chosen according to own preferences. 

At an upcoming intersection, risk is defined by the section of the path that lies within the junction.
Since risk corresponds to exposition time, we encode the path part from the intersection $I_z$ with a color, ranging from green for short intersections to red for long ones. 
Fig. \ref{fig:charts}~a) gives two examples of the hazard route.
The left bar shows a large intersection (e.g. multi-lane four-way stop) in vicinity and the right bar has a small and consecutive medium junction. 
This emphasizes that we may include more than one intersection in our warnings.

\subsubsection{Velocity Scale Element}
The velocity scale, Fig. \ref{fig:charts}~b), is a second chart element which qualifies the difference between the current velocity of the vehicle $v_0$ and the target velocity $v_{\text{tar}}$ from the trajectory evaluation of section \ref{subsec:trajeval}. 
The scale shows possible velocity values, from standstill $v\hspace{-0.05cm}=\hspace{-0.05cm}\unit[0]{m/s}$ to a maximal velocity $v_{\text{max}}$. Depending on the difference $|v_0 \hspace{0.05cm} - \hspace{0.05cm} v_{\text{tar}}|$, the situation is rated as safe with $v_0 \hspace{-0.042cm} \approx \hspace{-0.042cm} v_{\text{tar}}$ (green, left), as dangerous with e.g. $v_0 \hspace{-0.05cm} < \hspace{-0.05cm} v_{\text{tar}}$ (yellow, middle) to critical with $v_0 \hspace{-0.07cm} \ll \hspace{-0.07cm} v_{\text{tar}}$ (red, right). The same cases hold true for the opposite circumstances, i.e., $v_0 \hspace{-0.032cm} > \hspace{-0.032cm} v_{\text{tar}}$. 
This velocity scale can be employed for curve or regulatory risks. 
Moreover, we may set an enforced speed limit as the target velocity $v_{\text{tar}}$ for proactive behavior, once there is no risk ahead. 

\subsection{Short-Term Warning Elements}
\label{subsec:warning}
In order to emphasize the criticality of the situation, we propose to add further intuitive warning elements as e.g. pop-up signs and lane colorings. 
The following elements augment the proactive elements.

\subsubsection{Pop-up Signs}
Explicit symbols indicate the risk cause accompanied with the event time for collisions ($s_E$), distances to the risk spot for turns (i.e., right curve with $d_r$ and left curve with $d_l$) or stopping distance for crosswalks ($d_c$). In Fig. \ref{fig:popups}~a), the pop-up signs are pictured. 
We want to stress that this is just a selection and more risk causes can be added. 
The purpose is also to clarify the reason for the warning and give more human-understandable information.

\subsubsection{Colored Events}
\begin{figure}[t]
  \centering
  \resizebox{\linewidth}{!}{
  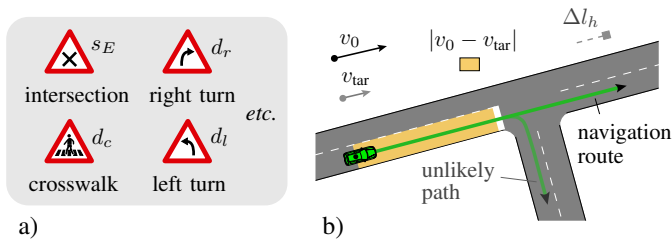}  
  \vspace{-0.53cm}
  \caption{Short-term warning elements. Selected pop-up warnings (left) and colored lane (right).}
  \label{fig:popups} 
\end{figure}

Finally, we highlight lane parts or positions according to the corresponding risks.  
In the instance of curve and regulatory risk, the lane is colored from the ego position up to the point of maximal risk. 
For collision risk, we mark the point of the closest encounter as a red cube.
An illustration for regulatory risk induced from a stop line is depicted in Fig. \ref{fig:popups}~b). Again, the color is defined by the deviation $|v_0-v_{\text{tar}}|$. It also shows the therein considered navigation route with length $\Delta l_h$ and another unlikely path. 

It should be noted that the visualization of warnings only occurs if the risks are actually present. 
Altogether, the RNS provides a variety of tools to analyze and circumvent critical situations in intersection scenarios, while not overloading the driver's awareness.

\section{Field Tests}
\label{sec:fite}
This section presents an evaluation of the complete Risk Navigation System (RNS) from Fig.~\ref{fig:blockdiagram}. 
We tested RNS offline on real-world recordings using the middleware RTMaps.\footnote{For the RTMaps software, please consult www.intempora.com.}
The mobile platform is a modified Honda CR-V, equipped with an OXTS localization device\footnote{Specifications of OXTS gear are described in www.oxts.com.}  
and six Ibeo Lux lidars allowing $\unit[360]{^\circ}$ perception.\footnote{At last, details of Ibeo sensors can be found on www.ibeo-as.com.} 
Additionally for debugging, a front-facing camera provides images. 
In the current implementation, RNS runs with a frequency of $\unit[10]{Hz}$ while employing the library OpenGL for environment rendering.

\begin{figure*}
  \centering
  \begin{overpic}[width=0.9853\linewidth]{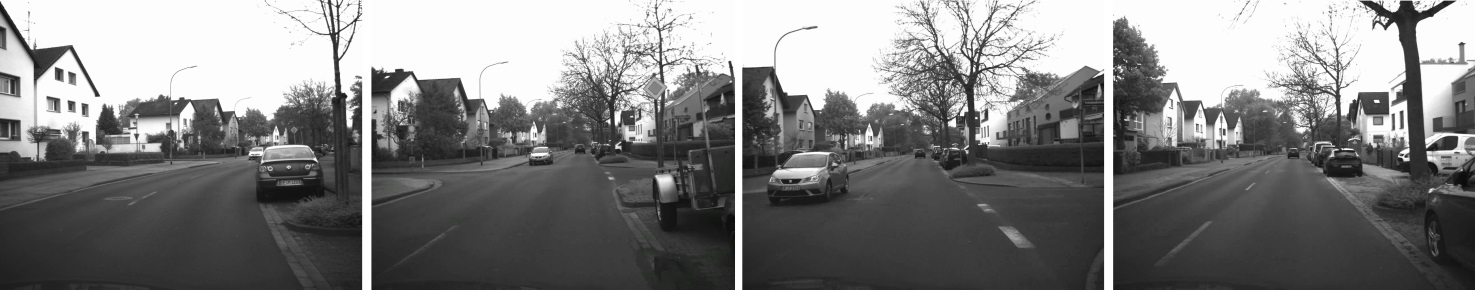}
    \put(24.5,-1){\color{white}\rule[0pt]{3.63pt}{110pt}}
    \put(49.675,-1){\color{white}\rule[0pt]{3.48pt}{110pt}}
    \put(74.85,-1){\color{white}\rule[0pt]{3.42pt}{110pt}}
  \end{overpic}
  \resizebox{0.24135\linewidth}{!}{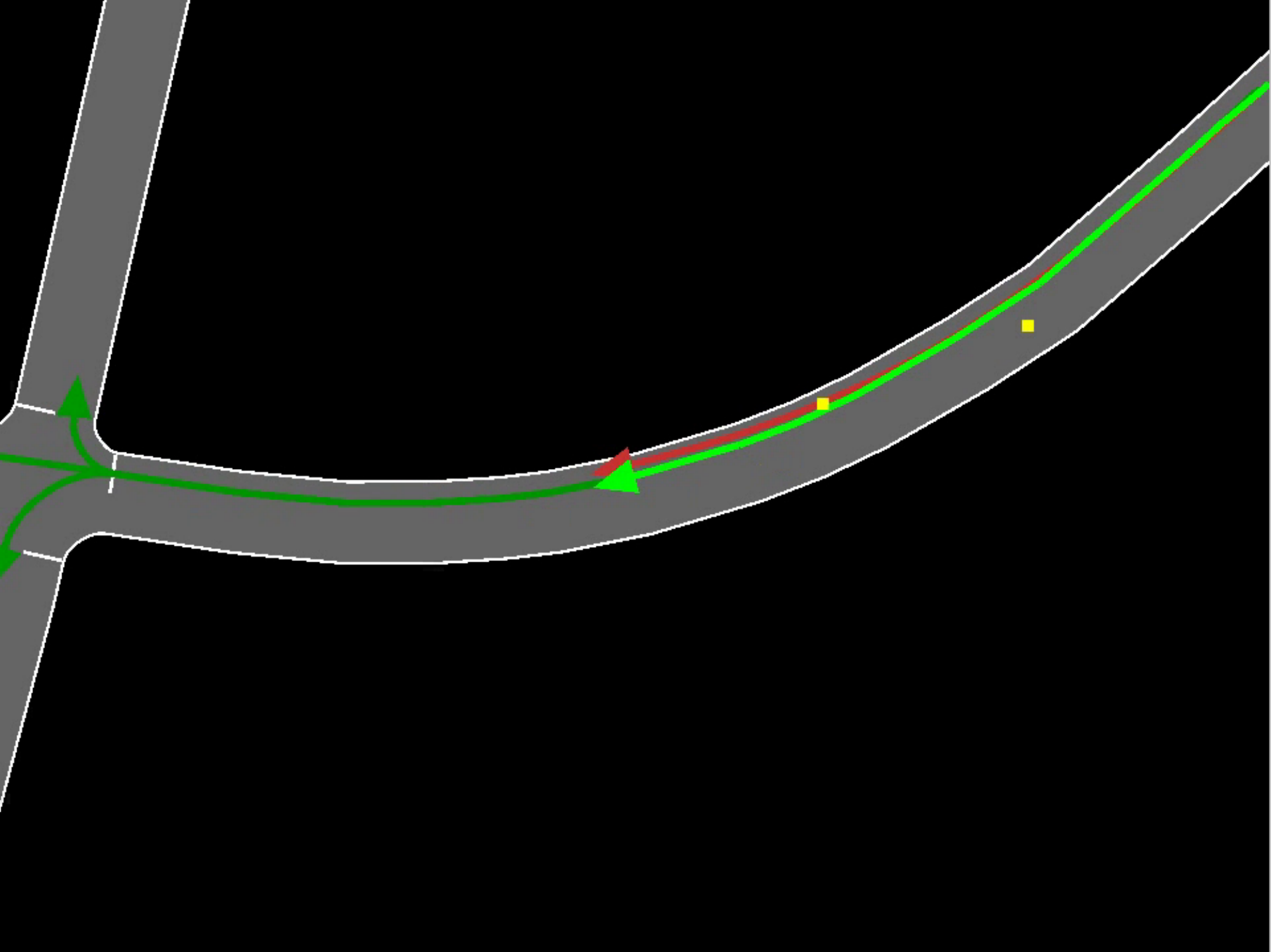}
  \resizebox{0.2411\linewidth}{!}{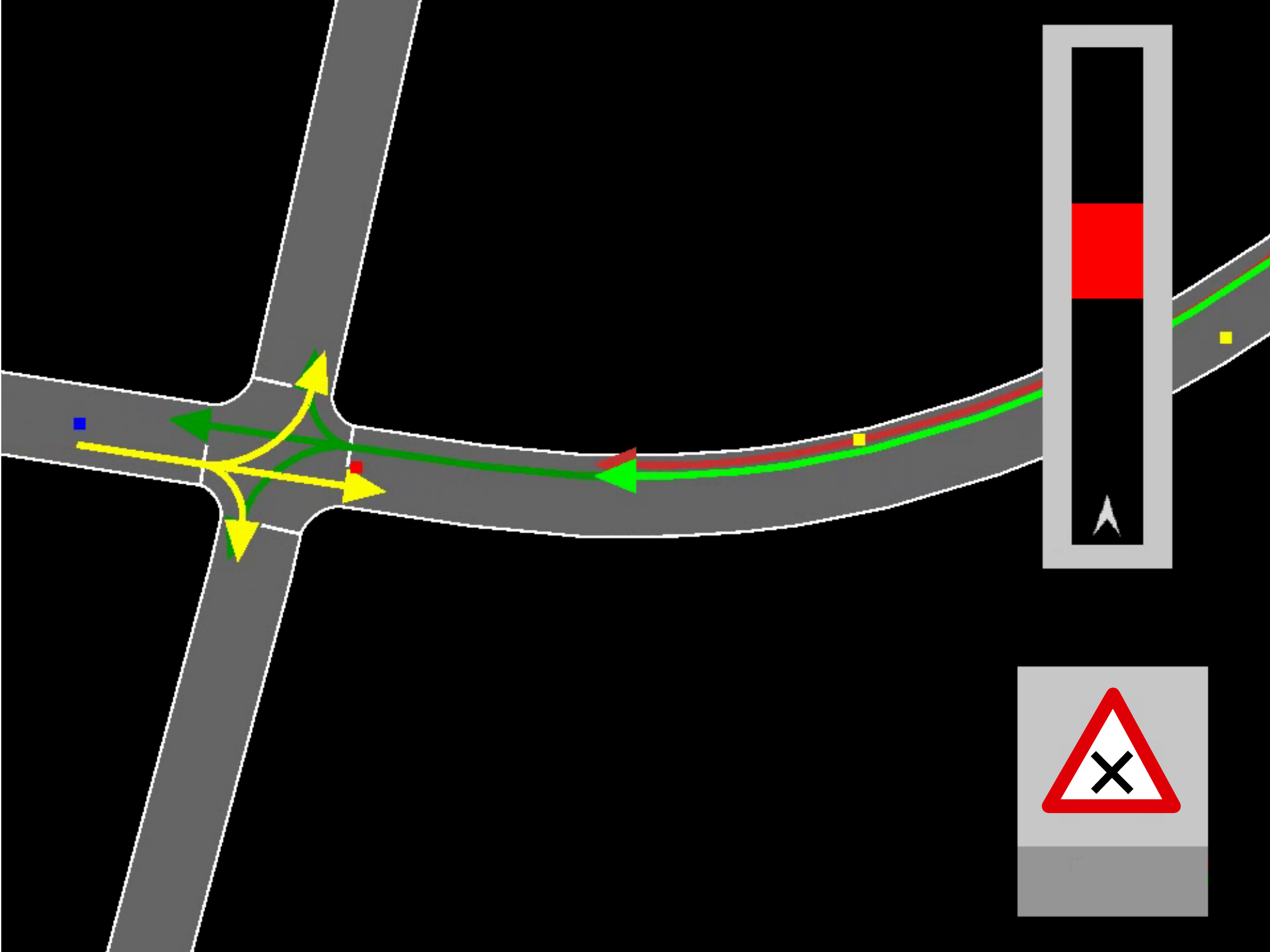}
  \resizebox{0.2411\linewidth}{!}{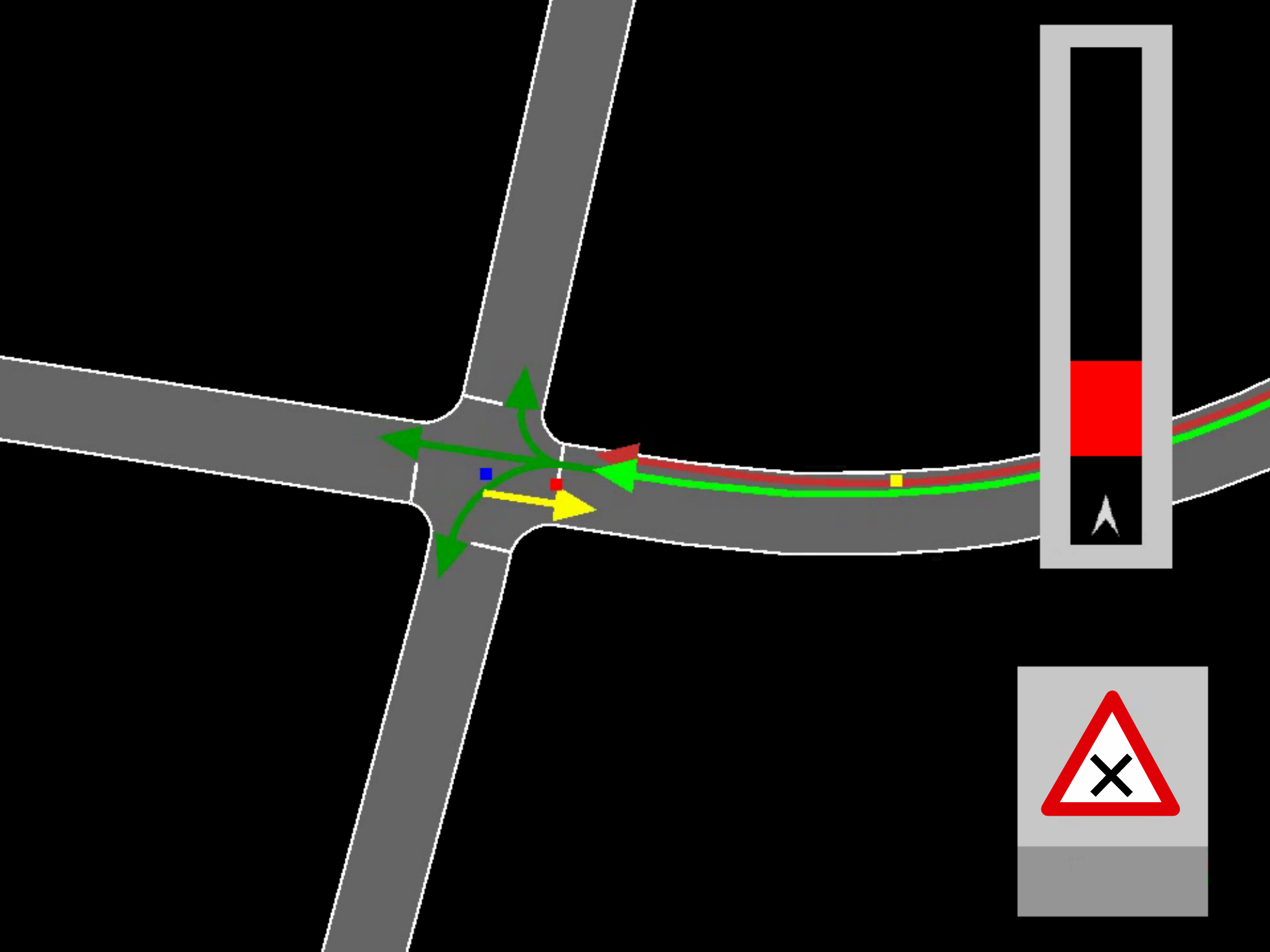}
  \resizebox{0.24085\linewidth}{!}{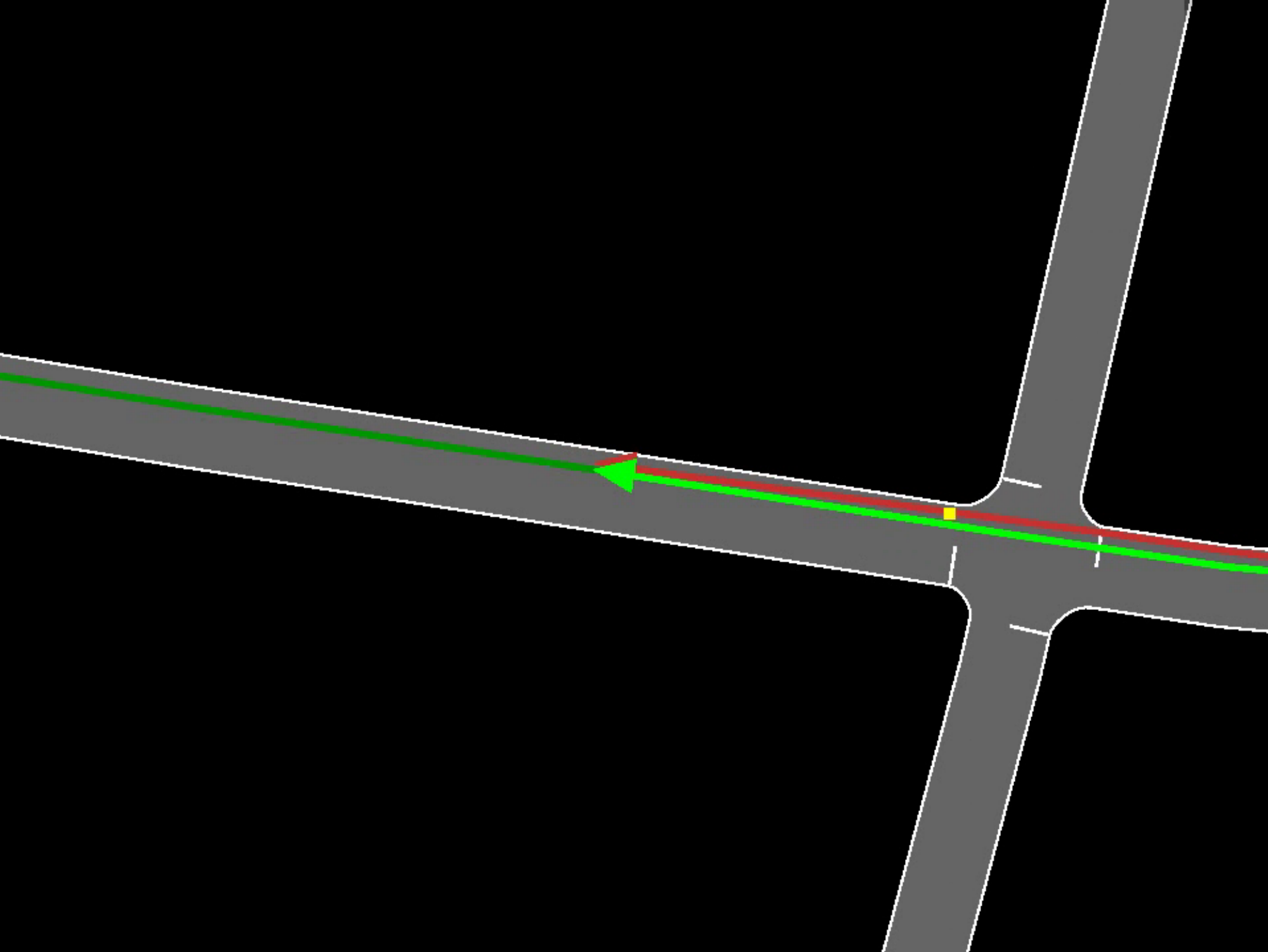}
  \caption{Collision risk example. At intersections, RNS calculates TTCE for combinations of ego plus other vehicle's path and informs about the critical encounter point (red line: recorded trace from ego car, green line: trace projected on navigation path). Top: Stream of front camera images. Bottom: Screen layout of system.}
  \label{fig:scenario1}
\end{figure*} 

\begin{figure*}
  \centering
  \begin{overpic}[width=0.981\linewidth]{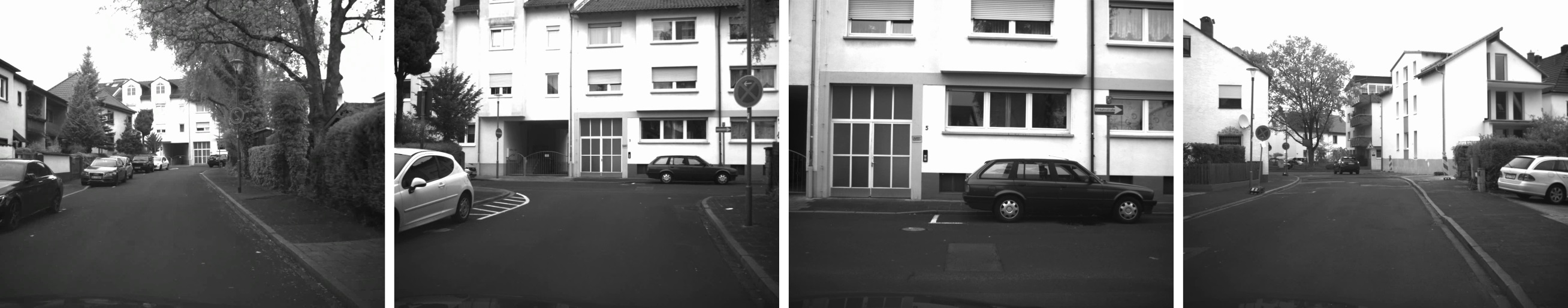}
    \put(24.49,-1){\color{white}\rule[0pt]{3.4pt}{110pt}}
    \put(49.65,-1){\color{white}\rule[0pt]{3.48pt}{103pt}}
    \put(74.83,-1){\color{white}\rule[0pt]{3.42pt}{110pt}}
  \end{overpic} \\
  \vspace{-0.015cm}
  \resizebox{0.24\linewidth}{!}{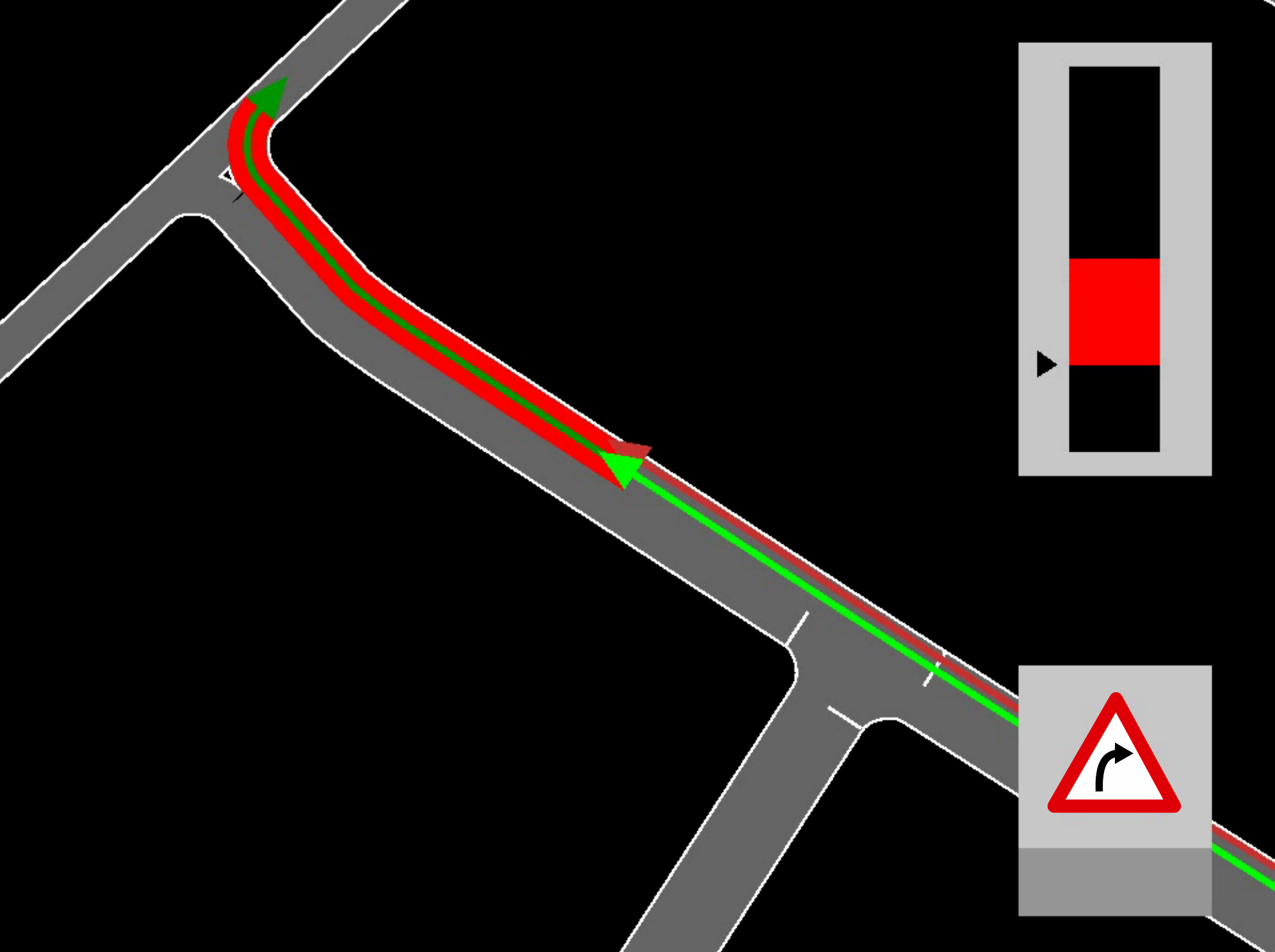}
  \resizebox{0.24\linewidth}{!}{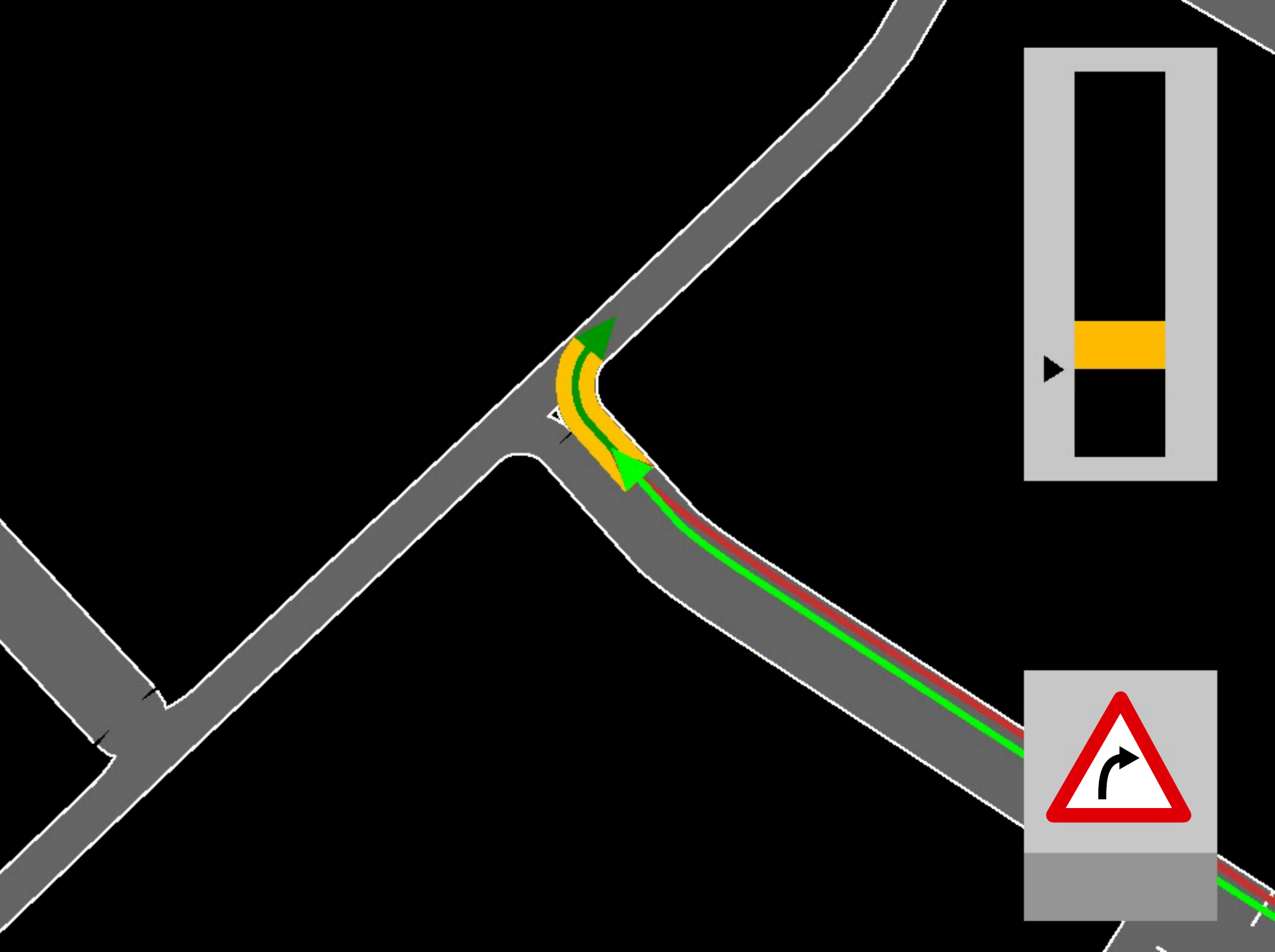}
  \resizebox{0.24\linewidth}{!}{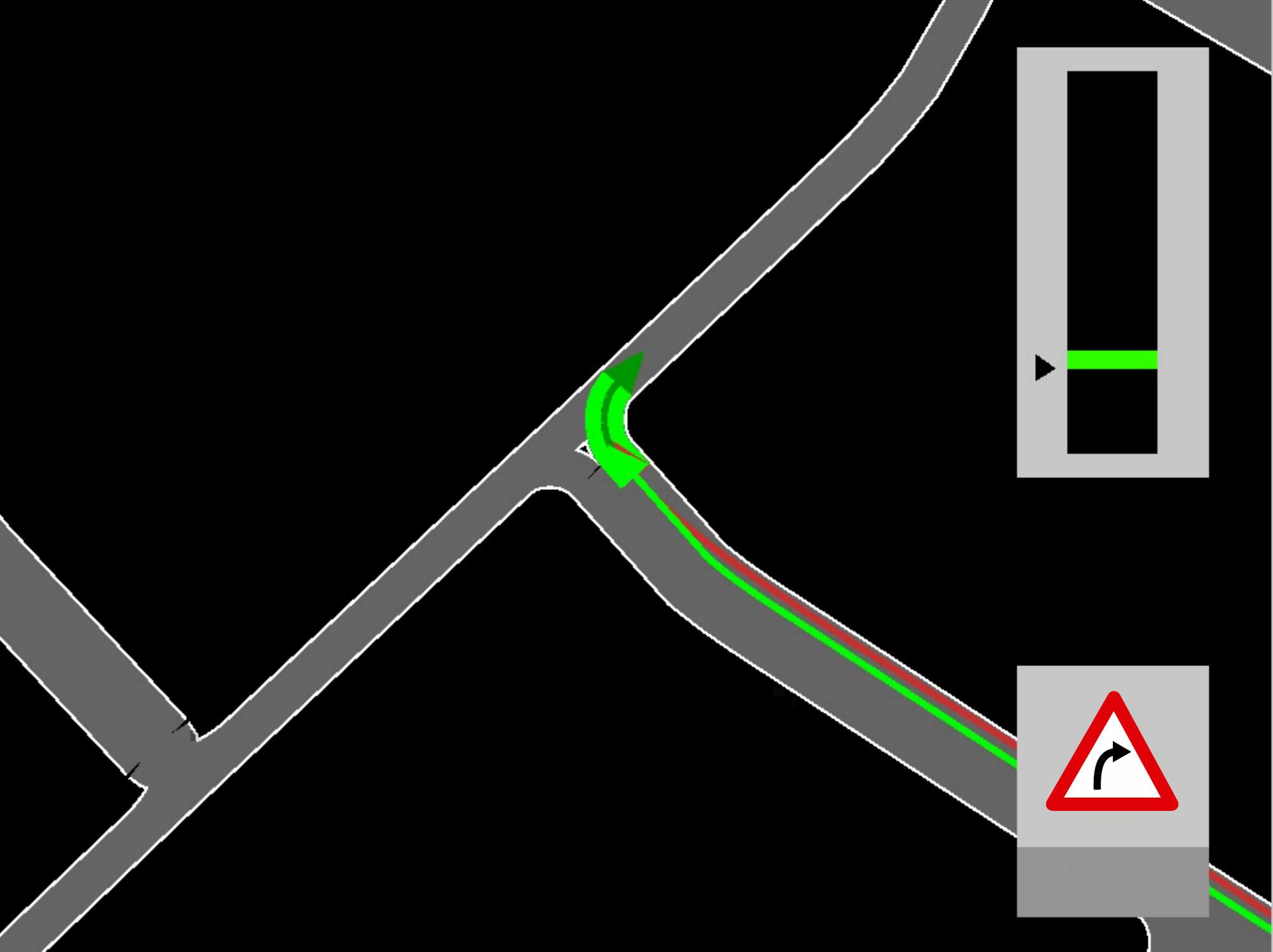}
  \resizebox{0.24\linewidth}{!}{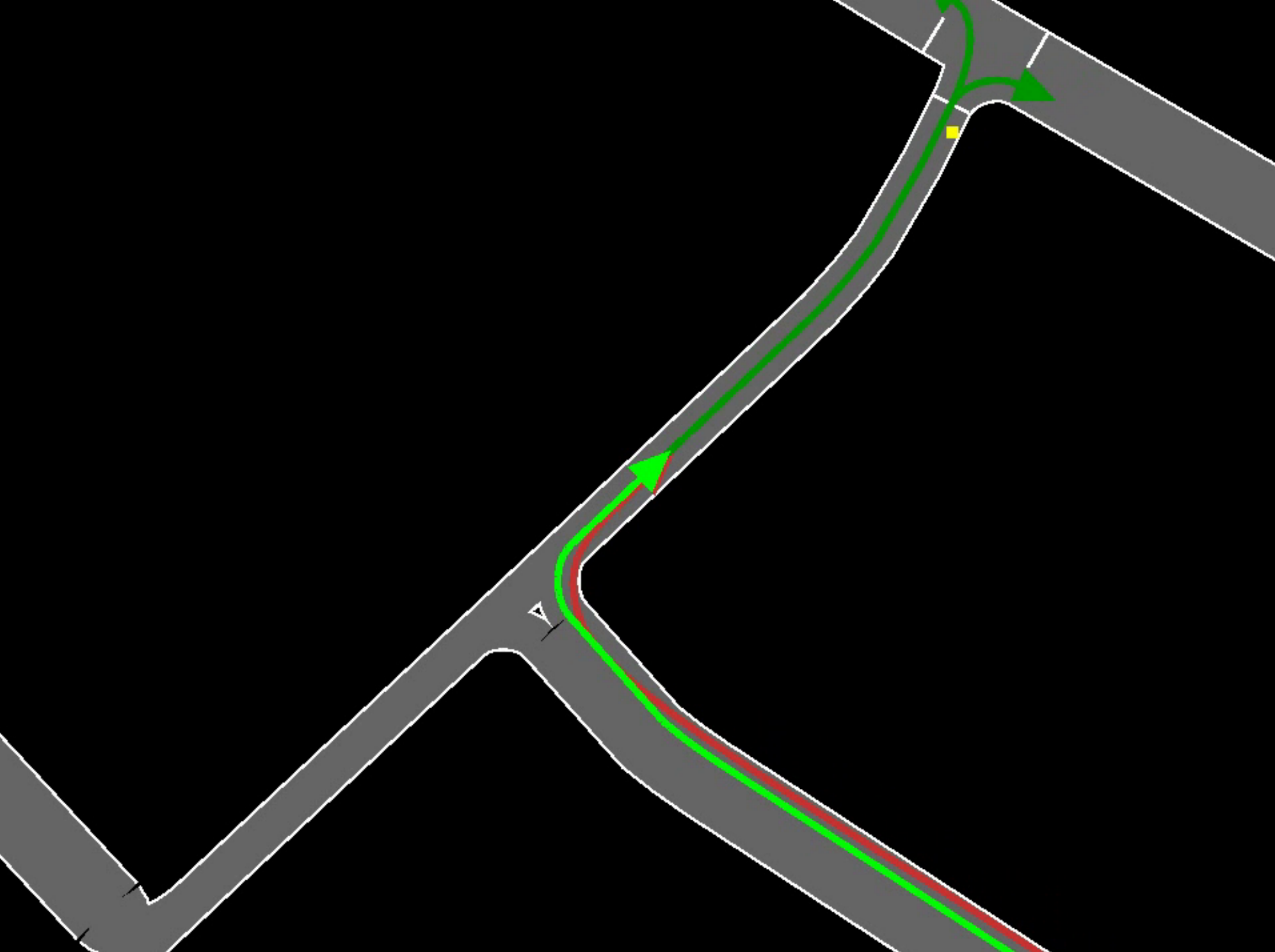}
  \caption{Curve risk example. RNS detects a sharp turn and recommends to decrease the velocity, which the driver is shortly after abiding to. The velocity scale goes from red (difference of actual to desired velocity) to green (small difference).}
  \label{fig:scenario2}
\end{figure*} 

\begin{figure*}
  \centering
  \begin{overpic}[width=0.9816\linewidth]{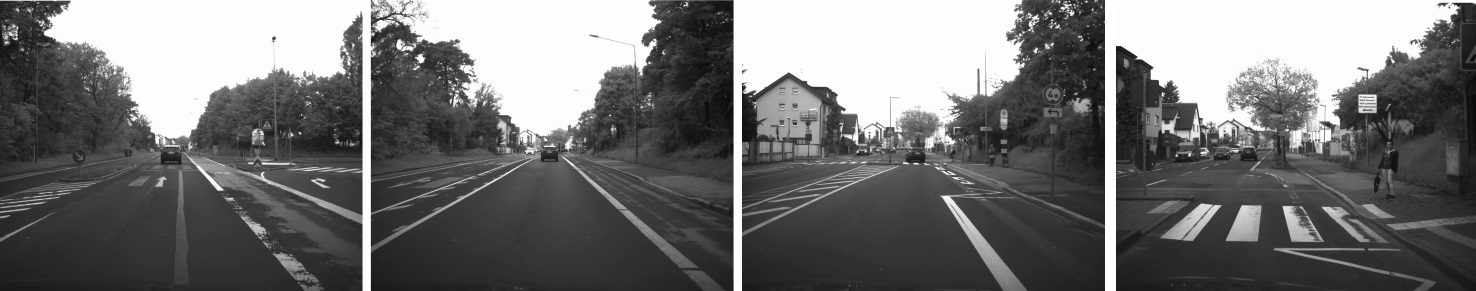}
    \put(24.49,-1){\color{white}\rule[0pt]{3.45pt}{110pt}}
    \put(49.635,-1){\color{white}\rule[0pt]{3.45pt}{110pt}}
    \put(74.81,-1){\color{white}\rule[0pt]{3.96pt}{110pt}}
  \end{overpic}
  \\
  \resizebox{0.24025\linewidth}{!}{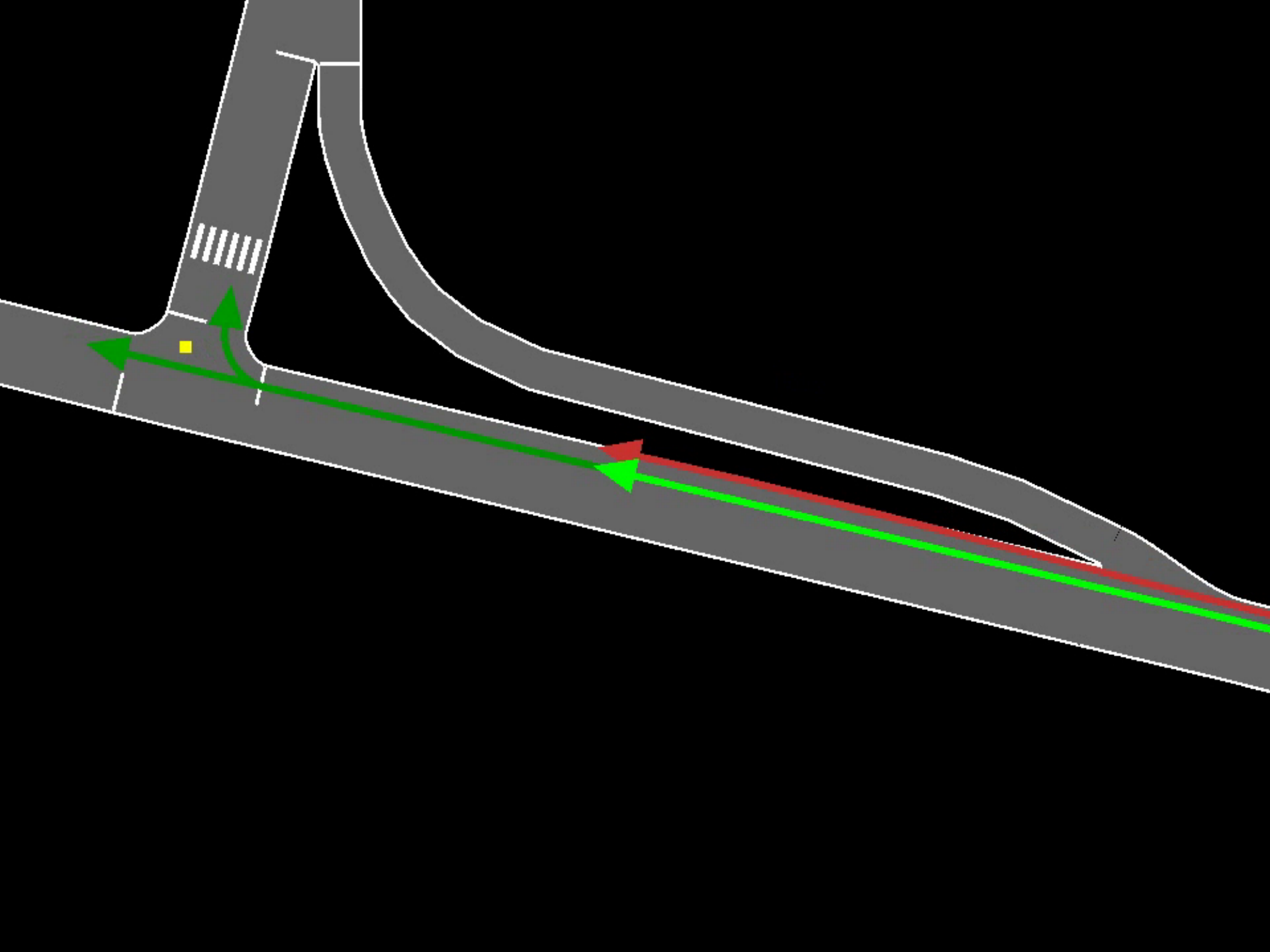}
  \resizebox{0.24\linewidth}{!}{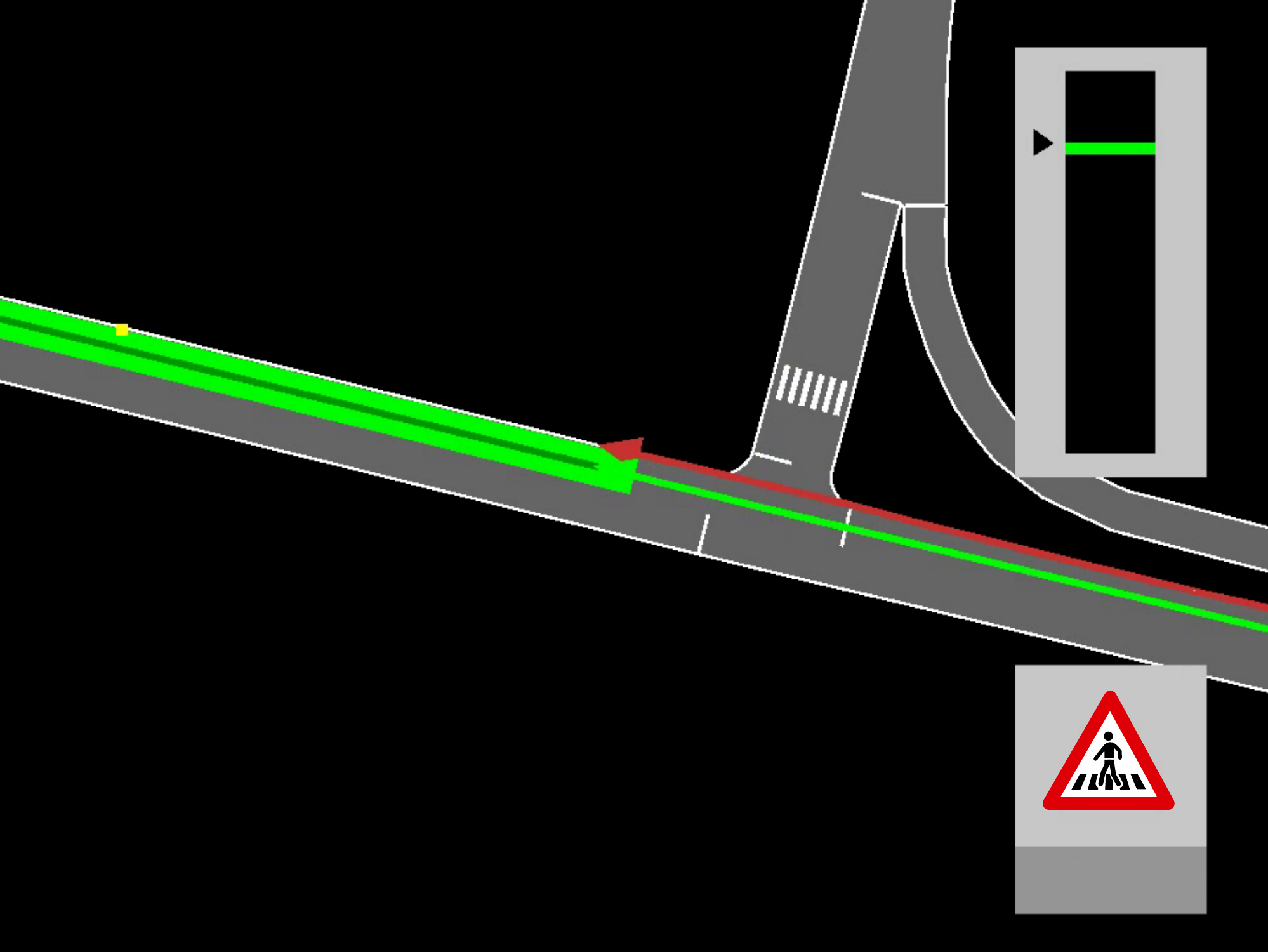}
  \resizebox{0.24025\linewidth}{!}{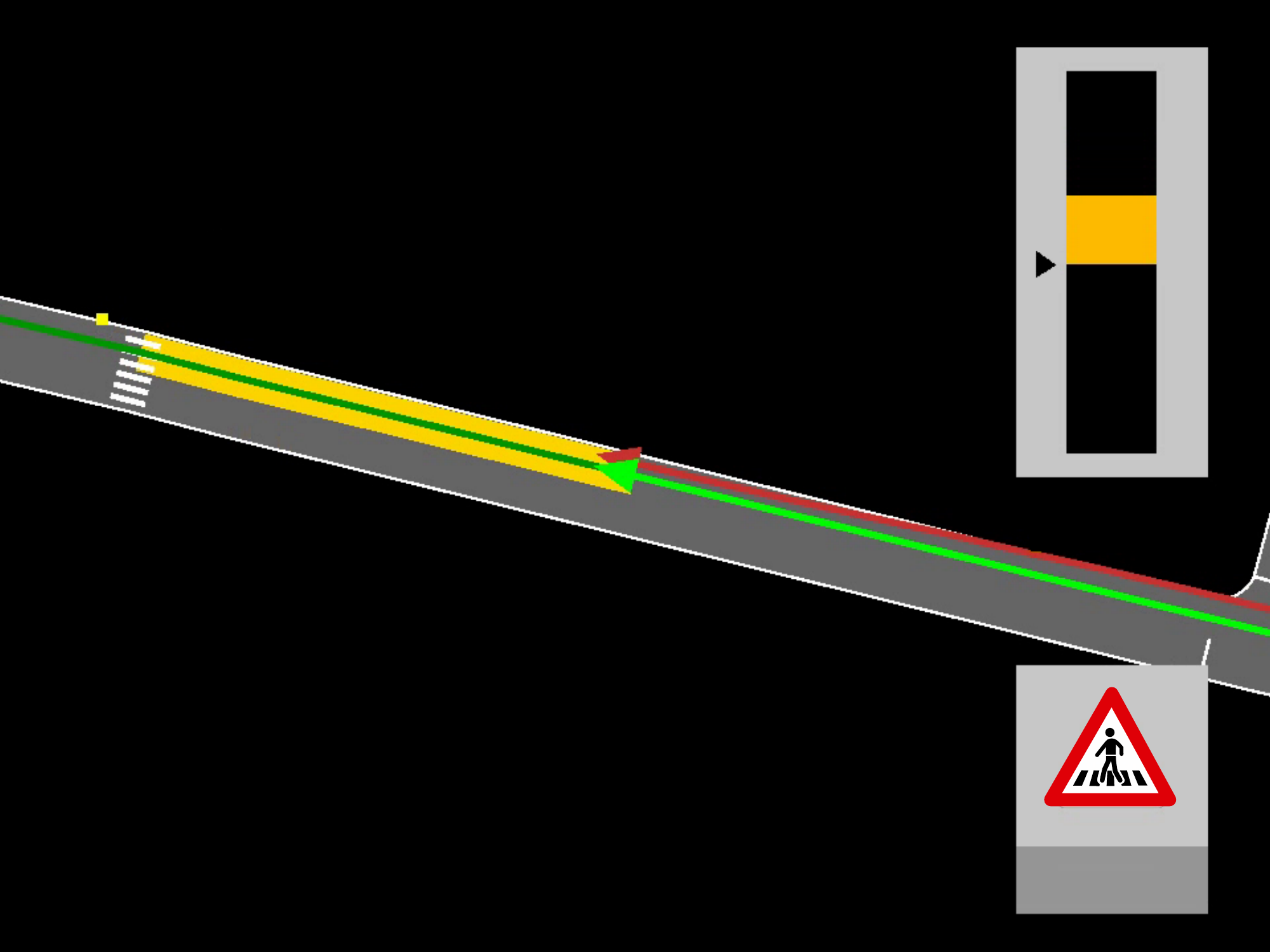}
  \resizebox{0.24025\linewidth}{!}{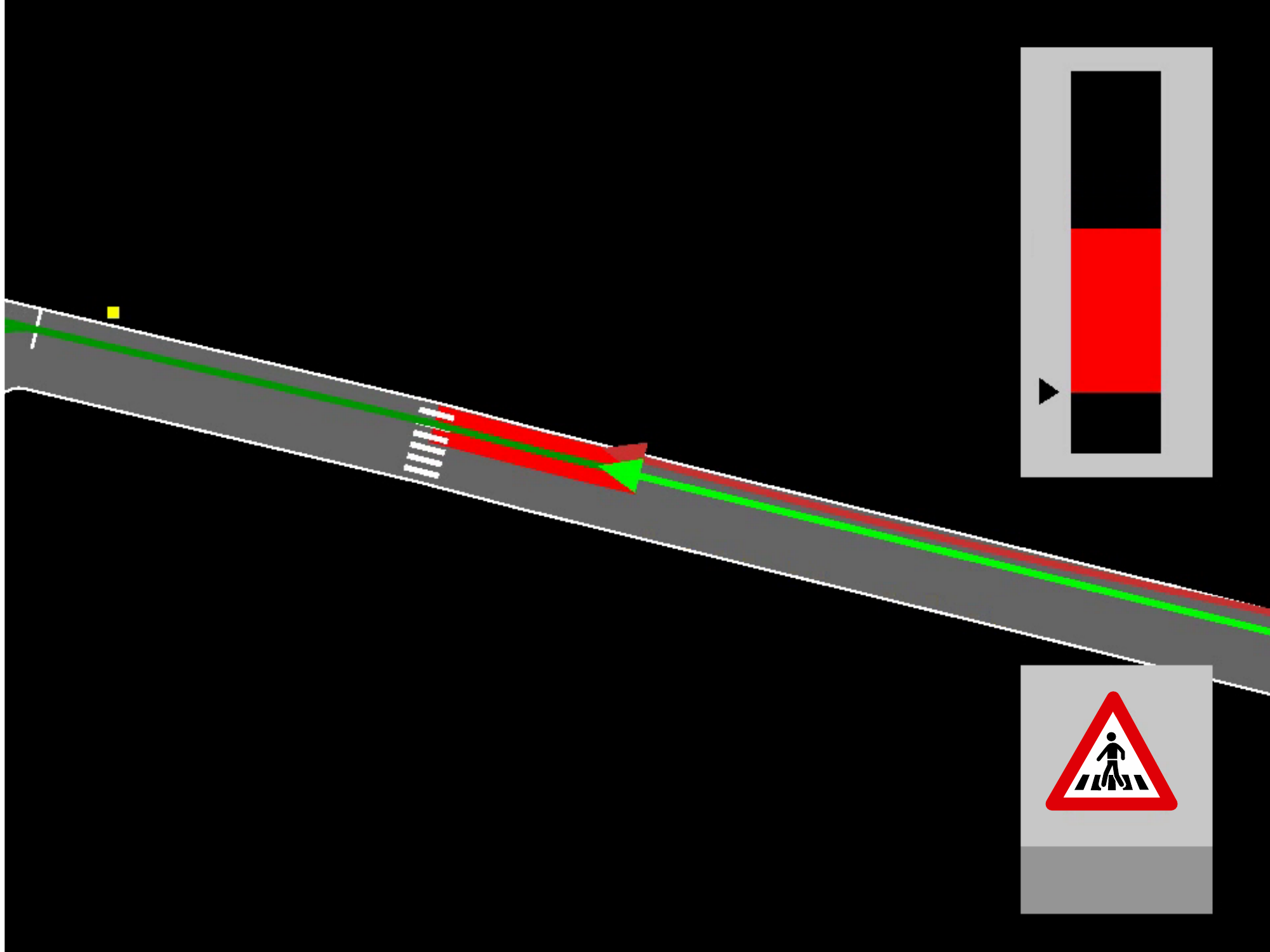}
  \caption{Regulatory risk example. In the situation, the driver does not let a prioritized pedestrian pass first on the crosswalk. RNS warns of the upcoming traffic object and displays the distance and colors the ego lane according to the risk.}
  \label{fig:scenario3}
\end{figure*} 

Multiple recordings have been acquired in Offenbach am Main, Germany. 
Necessary inputs of the system are positions, velocity estimates and angle measurements of both ego car and surrounding vehicles. 
Regarding localization of the ego vehicle, we project the GNSS signal onto its navigation route. 
Therefore, we have to know the route of the user in advance.
Alternatively, a filter has to be added that selects the most likely ego path.  
Throughout the course of the experiments, the R-LDM is queried for upcoming risks-related data around the current position. 

\subsection{Qualitative Outputs}
\label{sub:quali}
In three intersection cases, we assess the support of RNS for \textit{approaching}, \textit{turning} and \textit{crossing} tasks. 
Qualitative outputs of RNS visualizations are shown. 
Furthermore, we provide values of distances $d_I$, $d_c$, $d_r$ and times $s_E$ as well as driven $v_0$ and target $v_{\text{tar}}$ velocity. 
If needed, the quantitative values can be turned off (slim mode) in order to reduce the workload for the user. 

The first test run is an X-shaped junction with one oncoming other car. 
Four chronologically sorted scenes are given in Fig. \ref{fig:scenario1} with front camera images on top and the synchronized RNS output on the bottom. 
We show the recorded ego trace with a red line and projected position as a green arrow tip.
For the prediction horizon, we set $\Delta l_h = \unit[50]{m}$. 
Other detected cars are visualized as yellow dots, while the most critical other vehicle is depicted in blue with its possible paths being displayed as well. 

In the initial image, the ego car is approaching the intersection. 
As can be seen, the junction lies in a distance of $\unit[30]{m}$ for the next picture. 
RNS shows the point of closest encounter and indicates the event time $s_E\hspace{-0.02cm}=\hspace{-0.02cm}\unit[3]{s}$ under the given traffic sign. 
At the time of the third image, we nearly passed the other vehicle with $s_E\hspace{-0.05cm}=\hspace{-0.05cm}\unit[1]{s}$ and are close to the intersection, i.e. distance $d_I\hspace{-0.06cm}=\hspace{-0.06cm}\unit[15]{m}$. 
Finally, no critical situation is present anymore in the last image.
The scenario shows how RNS can \textbf{inform} the driver about possible situations and improves the prediction capability.

The second test represents a T-junction with a sharp right turn (see Fig. \ref{fig:scenario2}). 
In this context, $v_{\text{tar}}\hspace{-0.03cm}=\hspace{-0.03cm} \unit[4]{m/s}$ describes the velocity the ego driver should have adopted when reaching the curve.
Since the vehicle exceeds $v_{\text{tar}}$, the velocity scale turns red. 
However, due to appropriate behavior of the driver, i.e., reduction of speed $v_0$, the scale changes to yellow on the second image from the left. 
Here, the distance to the turn $d_r$ decreased simultaneously from $\unit[40]{m}$ to $\unit[10]{m}$. 
When arriving at the curve, the driver now matches the RNS target velocity  $v_0\hspace{-0.04cm}=\hspace{-0.04cm}v_{\text{tar}}$, denoted in green. 
This example shows how users can leverage \textbf{recommendations} from curve risks using the velocity scale of the RNS.

In the third experiment, a pedestrian intends to use a cross-walk and the car driver ignores the regulatory risk. 
It should be considered that this crosswalk does not have traffic lights, as it is common e.g. in Germany. Since the R-LDM stores traffic elements, we can warn the driver already $d_c=\unit[90]{m}$ in advance with a pop-up symbol. In the following sequence of Fig. \ref{fig:scenario3}, the driver keeps its velocity $v_0$ nearly constant while the suggested stopping trajectory $v_{\text{tar}}$ changes from $\unit[12]{m/s}$ with a green warning, via $\unit[7]{m/s}$ with a yellow velocity scale to $\unit[3]{m/s}$ at $\unit[10]{m}$ distance to the waiting pedestrian. Because of the deviation between the driver and RNS, the \textbf{warning} became critically red in the end. In turn, we are able to guide the driver's awareness towards the most relevant risks.

\subsection{Performance Discussion}
\label{sub:dicuss}
We showed the general applicability of our concept and focused on urban scenarios with single-lane streets.
When handling complex roads, accurate lane-level localization becomes more important.
On one side, the ego vehicle has to be projected on the right lane, which puts demands on the lateral localization precision. 
On the other side, obstacles that are sensed relative to the ego vehicle need to be assigned to their own proper lane, which adds further requirements to both lateral error and ego orientation estimation. 
To tackle this, we refer to e.g. \cite{flade2017}.

The results underline the potential of RNS to proactively support the driver. 
Its visual display allows us to consistently analyze the situational points of interests. In this sense, we provide a navigation on strategical and manoevring level with time horizons of several seconds.

\section{Conclusion and Outlook}
\label{sec:outl}
In this paper, we outlined a map-based risk warning device for urban driving. RNS aligns the GNSS position in a relational local dynamic map. Upcoming traffic objects are then queried along the navigation route (i.e., detected other cars from lidar, sharp turns and crosswalks). Here, we derive road geometries and object positions with augmented OpenStreetMap data and utilize direct data links in a graph representation. The situation can be evaluated with TTCE for car-to-car collision risk, maximal curvature for curve risk and the distance to full stop for regulatory risks. 

A detailed driving context is now visualized using a 2D/3D renderer. We chose OpenGL augmented by elements that convey the risk sources with spatio-temporal vehicle interactions. Each of them have the same underlying color code from green as safe to red as dangerous. With an application of RNS on real intersections, we eventually showed the versatile warning functionality. The user is supported to perceive the relevant obstacles in long range, safe behaviors are recommended in medium terms and risky encounters are highlighted for short times.    

Instead of pure GNSS usage, the RNS could benefit from adding vision-based localization components (as mentioned before).  
The visualizations of RNS depend on the quality of the map and also on the reference within the map. Matching detected lanes on images with stored street polylines is promising \cite{flade2018}. This potentially also reduce the errors from the shared map. Intersections are still to be investigated because of typically mixed curvy and straight segments. 

The risk prediction in RNS draws upon time metrics and does not consider uncertainties in driving. In previous research, we developed the survival analysis \cite{puphal2019} that consists of Gaussian and Poisson distributions for the modeling of states and critical events. It showed to detect risks early and with less false positives. Hence, we may be able to make RNS more effective. A combination of intuitive with uncertainty-aware risks for warning is possible. 

Furthermore, augmented reality is deployable besides the 2D screen display. A 3D environment could be simulated in first-person-view and overlayed on camera images. Here, the pitch angle vibration from bumps in the road has to be compensated by changing the camera angle of the simulator. We could thus circumvent the higher effort of processing 3D compared to 2D information.

Finally, the paper evaluated and discussed technological aspects of a Risk Navigation System and showed its potential. What remains to be done as a next clear step is a user study that investigates the ergonomics, psychological workload and societal acceptance.

\section*{Acknowledgment}
\noindent This work has been supported by the European Unions Horizon 2020 project \textit{VI-DAS}, under the grant agreement number 690772. 
Map data \copyright~OpenStreetMap contributors, licensed under the Open Database License. 
We like to thank Cristhiam Felipe Pulido Riveros from National University of Colombia for his support. 
\vspace{0.15cm}

\bibliographystyle{IEEEtran}
\bibliography{bib}

\end{document}